\documentclass{article}

\usepackage{microtype}
\usepackage{graphicx}
\usepackage{subcaption}
\usepackage{booktabs} 

\usepackage{hyperref}

\usepackage[preprint]{icml2026}

\usepackage{amsmath}
\usepackage{amssymb}
\usepackage{mathtools}
\usepackage{amsthm}

\usepackage[capitalize,noabbrev]{cleveref}

\theoremstyle{plain}

\theoremstyle{definition}

\theoremstyle{remark}

\usepackage[textsize=tiny]{todonotes}

\icmltitlerunning{PriorProbe for Neural Network Personalization}

\begin{document}

\twocolumn[
  \icmltitle{PriorProbe: Recovering Individual-Level Priors for Personalizing Neural Networks in Facial Expression Recognition}


  \icmlsetsymbol{equal}{*}

  \begin{icmlauthorlist}
    \icmlauthor{Haijiang Yan}{warpsy}
    \icmlauthor{Nick Chater}{warwbs}
    \icmlauthor{Adam Sanborn}{warpsy}
  \end{icmlauthorlist}

  \icmlaffiliation{warpsy}{Department of Psychology, University of Warwick, Coventry, United Kingdom}
  \icmlaffiliation{warwbs}{Warwick Business School, University of Warwick, Coventry, United Kingdom}

  \icmlcorrespondingauthor{Haijiang Yan}{haijiang.yan@warwick.ac.uk}

  \icmlkeywords{Machine Learning, ICML}

  \vskip 0.3in
]



\printAffiliationsAndNotice{}  

\begin{abstract}
  Incorporating individual-level cognitive priors offers an important route to personalizing neural networks, yet accurately eliciting such priors remains challenging: existing methods either fail to uniquely identify them or introduce systematic biases. Here, we introduce PriorProbe, a novel elicitation approach grounded in Markov Chain Monte Carlo with People that recovers fine-grained, individual-specific priors. Focusing on a facial expression recognition task, we apply PriorProbe to individual participants and test whether integrating the recovered priors with a state-of-the-art neural network improves its ability to predict an individual's classification on ambiguous stimuli. The PriorProbe-derived priors yield substantial performance gains, outperforming both the neural network alone and alternative sources of priors, while preserving the network’s inference on ground-truth labels. Together, these results demonstrate that PriorProbe provides a general and interpretable framework for personalizing deep neural networks.
\end{abstract}

\section{Introduction}

Adapting artificial intelligence to individual users is a critical step toward developing digital systems that deliver genuinely personalized assistance in everyday decision-making. This requires artificial intelligence to infer how users would respond to a given cue, but although modern neural networks have demonstrated substantial capacity to predict human behavior across tasks \cite{tan_artificial_1996, lin_predicting_2022, rafiei_neural_2024, Binz2025ACognition}, capturing individual behavior remains a significant challenge. Even the state-of-the-art deep neural networks for facial expression recognition struggle to capture individual-specific perceptual judgments, particularly on subtle expressions where substantial interpersonal variability is observed \cite{palermo2013new}.

Formally, given a facial expression $f$, individuals must infer the underlying emotional state $e$ of the person. Bayesian inference provides a principled framework for characterizing how to make optimal inferences under uncertainty:
\begin{equation}
    P(e|f) \propto P(e)P(f|e)
\end{equation}
where $P(e)$ denotes an individual’s prior belief over emotional states in the absence of facial evidence, while the likelihood $P(f|e)$ quantifies the degree to which the observed facial expression supports a particular emotional state. When $f$ strongly favors a specific $e$ as in most everyday situations, the likelihood term $P(f|e)$ dominates the inference. In such cases, neural networks with knowledge of facial expression patterns can readily predict how humans perceive $f$ \cite{canal_survey_2022}. However, when $f$ is ambiguous, the likelihood becomes relatively uninformative and prior beliefs $P(e)$ exert a greater influence on the inference. Consequently, nuanced individual differences in $P(e)$ can cause neural networks whose prior representations are misaligned with those of individuals to fail at predicting how individuals interpret ambiguous stimuli. 

Previous research has attempted to endow neural networks with effective priors by training them on synthetic datasets, including ecological priors to which humans may adapt \cite{jagadish_meta-learning_2025, zhu_language_2025}, as well as cognitive priors derived from established cognitive models \cite{Bourgin2019CognitiveDecisions, McCoy2023ModelingNetworks}. However, when such approaches are used to inform neural networks with an individual's prior, they face two fundamental challenges:
\begin{itemize}
    \item Training neural networks on implicit priors within a dataset typically relies on population-level assumptions. When extending to individuals, collecting sufficiently large and representative datasets at the individual level is often costly and practically infeasible. 
    \item Due to sparsity and noisy nature of individual behavior, finetuning neural networks on such data risks overwriting or degrading the model’s learned representations, causing catastrophic forgetting.
\end{itemize}

In contrast to parameter-level personalization, empirical approaches from cognitive science aimed at eliciting individual-level cognitive priors provide a promising pathway for addressing these gaps. Explicitly identified priors can be incorporated directly into the inference space of neural networks without modifying their weights, thereby offering an interpretable means of altering network behavior with clear correspondence to cognitive constructs. However, identifying fine-grained individual priors comprehensively remains challenging. Generally, elicitation methods infer human priors from observed behavioral outputs \cite{kording_bayesian_2004, houlsby_cognitive_2013, yeung_identifying_2015}, while treating the likelihood function as either a free parameter or an arbitrarily specified mapping. As a consequence, such approaches may fail to uniquely identify priors or may introduce systematic biases in the recovered prior distributions.

\begin{figure}[ht]
  \vskip 0.2in
  \begin{center}
    \centerline{\includegraphics[width=\columnwidth, trim={9cm 6cm 9.5cm 8cm}, clip]{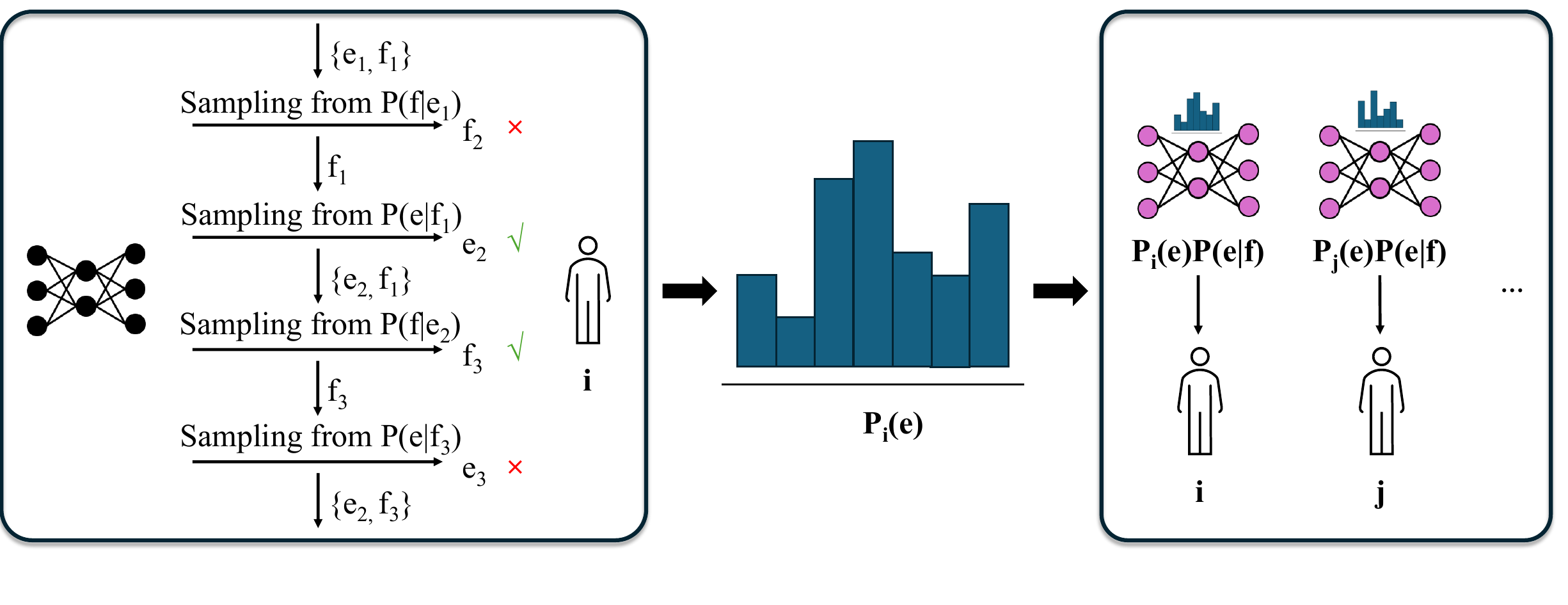}}
    \caption{
      The framework of using PriorProbe to inform neural network with individual-level preferences. The left panel illustrates a typical PriorProbe procedure, and the right panel describe how the recovered individual prior informs a shared neural network about individual differences in inference. 
    }
    \label{diagrame}
  \end{center}
\end{figure}

In this paper, we introduce a novel approach, PriorProbe, that recovers fine-grained, individual-specific prior representations while simultaneously identifying the corresponding individual-specific likelihood representations. The method is grounded in a block Metropolis–Hastings procedure, in which trials for hypothesis inference and trials for stimulus generation alternate within a single Markov chain with a human in the loop, enabling the joint collection of samples from both of this individual's representations (as shown in \cref{diagrame}). 

Focusing on a facial expression categorization task, our goal here is to recover individual-level cognitive priors over emotion categories and to examine whether integrating these priors with neural networks within inference space can improve their alignment with individuals. To this end, we first construct a controlled yet highly expressive sample space of facial expressions, DistFace, within which individual internal generative models for facial expressions can be elicited without bias. Within this space, PriorProbe is applied to individuals to recover their priors. Finally, by combining these priors with a state-of-the-art deep neural network, we evaluate whether this enables the model to better capture individual preferences without hurting the model's performance on unambiguous stimuli. Our contributions are summarized as follows:
\begin{itemize}
    \item We propose the novel PriorProbe for recovering fine-grained individual mental representations with high efficiency. 
    \item To facilitate PriorProbe in facial expression recognition, we construct DistFace for sample generation that disentangles facial expression with facial identify, providing a pure affect space for more accurate elicitation.
    \item We conduct a comprehensive human experiment to validate whether the state-of-the-art neural network informed by the recovered priors show better alignment with individual human, and importantly, without degradation of performance on ground-truth inference. 
\end{itemize}
    Overall, our work shows an explainable, general and robust framework to improve individual-specificity of neural network models.

\section{Related work}

\subsection{Personalization of Neural Networks}
Prior work on neural network personalization has primarily focused on recommendation systems, where models infer individual preferences from limited historical interactions \cite{im_case-based_2007, quadrana_personalizing_2017, nguyen_convolutional_2019, wang_dan_2020, wu_federated_2022}. With the recent rise of large language models and increasingly intensive human–AI interactions, personalization has been extended to more general domains beyond recommendation. In this context, latent conditioning approaches have been proposed to personalize language models by conditioning a shared network on individual-specific latent embeddings, thereby adapting model behavior without directly modifying network weights and reducing interference across users \cite{huber_embedding--prefix_2025, jaech_low-rank_2018, chen_when_2024}. Despite this progress, relatively little work has examined the personalization of neural networks for task-specific perceptual or cognitive modeling problems, such as facial expression recognition.

More broadly, existing approaches to neural network personalization typically come down to finetuning pretrained models for individual users. Standard fine-tuning methods are known to suffer from overfitting and catastrophic forgetting when individual-level data are sparse and noisy \cite{yosinski_how_2014, howard_universal_2018}. To alleviate these issues, parameter-efficient techniques (e.g., low-rank adaptations) restrict personalization to a small subset of parameters while freezing the shared backbone \cite{hu2022lora}. Meta-learning approaches further conceptualize personalization as rapid adaptation from few examples, often formalized within a hierarchical Bayesian framework that learns priors over task- or user-specific parameters \cite{finn_model-agnostic_2017, grant_recasting_2018}. 

\subsection{Human Priors Elicitation}
More directly, researchers can ask people to estimate the base rates of underlying hypotheses (e.g., emotion categories in our case), leveraging the close correspondence between optimal statistical inference and human cognitive judgments \cite{griffiths_optimal_2006}. However, such expressed probabilities are often incoherent and susceptible to substantial reporting biases \cite{tversky_extensional_1983, costello_surprisingly_2014, Zhu2020TheJudgments}. 

Model‑based inference instead relies on reverse Bayesian computation, which posits an underlying Bayesian model which behavior arises from and recovers those priors as well as other mental representations by fitting the model to observed behavioral data \cite{kording_bayesian_2004, houlsby_cognitive_2013}. However, a central limitation of this approach is that different combinations of priors and likelihoods might give rise to indistinguishable behavior, preventing their unique recovery from behavior alone. One way to mitigate this issue is to assume a fixed, universal likelihood and to design stimuli that amplify the influence of participants’ priors (e.g., through ambiguity or noise). 

Stemming from Markov Chain Monte Carlo (MCMC) algorithms, sampling-based elicitation circumvents the reverse inference by directly sampling from the target mental representations. Iterated learning (IL), for example, asks participants to perform inductive inference on each trial given a stimulus, and then generate the stimulus for the subsequent trial conditioned on the inferred hypothesis \cite{yeung_identifying_2015, zhu_eliciting_2024}. The chained inferences instantiate a Gibbs sampling process whose stationary distribution corresponds to the individual’s prior. However, iterated learning similarly imposes a ground‑truth likelihood function to generate stimuli and drive the sampling process, and this likelihood may be misaligned with individuals’ true category distributions, potentially biasing the stationary distribution of the resulting MCMC process. By contrast, Markov Chain Monte Carlo with People and Gatekeepers (MCMCPG) by Yan et al. \yrcite{yan_quickly_2024} draws samples from each individual’s category distribution (i.e., likelihood in this context) based on a Metropolis-Hasting process. However, MCMCPG does not extend to the recovery of individual priors. Overall, there remains a lack of elicitation approaches capable of jointly and uniquely identifying both individual-specific priors and likelihood representations for a simple inference task.

\section{DistFace: Disentangling Affect from Identity in the Face Generative Space}

PriorProbe samples from human mental representations by repeatedly eliciting choices over randomly generated stimuli and candidate hypotheses. As a result, defining a stimulus space that is both compact and sufficiently expressive for the target task is a critical basis for enabling efficient and informative sampling. In our case of facial expression, we construct a purely affect‑based representational space disentangled from facial identity by tailoring a Variational Auto‑Encoder (VAE) of face generation that allocates facial affect and identity to distinct sub‑latent spaces.

\subsection{Model Architecture}
The model comprises three components: (1) a hybrid encoder module $E$ composed of pretrained models that separately extract identity and affect representations; (2) a decoder $D$ that reconstructs faces from the hybrid representational space; and (3) an separately trained denoising auto‑encoder $R$ that further compresses and regularizes this space, providing a low-dimensional sampling space for PriorProbe. The architecture is shown in Appendix  \cref{DistFace}.

\subsubsection{Encoder module}
The first part of the hybrid encoder module consists of three sophisticated pretrained models that extract representations of facial expression, facial action unit (AU), and identity from given faces, respectively. For facial expression encoder $E_{fe}$, we adopted ResEmoteNet from Roy et al. \yrcite{roy2024resemotenetbridgingaccuracyloss}, which is a residual network trained on FER2013 dataset for facial expression recognition. ResEmoteNet achieved a SOTA performace on FER2013 with a test accuracy of 77.88\%. To complement facial expression features with a more granular representation of facial affect, we used the Facial Masked Autoencoder with Identity Adversarial Training (FMAE‑IAT) developed by Ning et al. \yrcite{ning2024representation} as another encoder branch $E_{au}$ to extract AU representations. AU encodes individual muscle movements of complex facial expressions, providing a rich set of expressive features for facial affect. The FMAE‑IAT builds on a MAE‑pretrained Vision Transformer (ViT) and was finetuned on the BP4D dataset \cite{zhang_bp4d-spontaneous_2014} for the AU detection task while explicitly maximizing an identity‑adversarial loss. As a result, the FMAE‑IAT yields affective features that are effectively independent of facial identity. For facial identity branch $E_{id}$, we used the Arcface, a deep residual network trained on Glint360k \cite{an_2022_pfc_cvpr, an_2021_pfc_iccvw} for face recognition, to yield features of face identity \cite{deng2019arcface}. For all three pretrained models, the final output layers for different tasks were removed to extract their hidden representations. Eventually, for a face input $f$, these models produce $E_{fe}(f) \in R^{1024}$, $E_{au}(f) \in R^{1024}$, $E_{id}(f) \in R^{512}$ in parallel, projecting facial affect and identity into distinct spaces. During the following training session, the weights of these pretrained models were fixed. 

The second component of the encoder module consists of two trainable multilayer perceptrons (MLP), $E^r_{af}$ and $E^r_{id}$. One receives the facial affect features formed by concatenating the output of $E_{fe}$ and $E_{au}$ (the affect branch), while the other receives output of $E_{id}$ (the identity branch). Each encoder maps its respective input into a dedicated 128‑dimensional latent space: 
\begin{equation}
    \{V_{af}, V_{id}\} = \{E^r_{af}([E_{fe}(f), E_{au}(f)]), E^r_{id}(E_{id}(f))\}
\end{equation}

\subsubsection{Decoder module}
The resulted 256-dimensional representation $[V_{af}, V_{id}]$ is then passed through the decoder of the VAE for image reconstruction following reparameterization. The decoder is implemented as a ResNet-style convolutional network with three residual blocks followed by five up-sampling layers. The facial representation is first projected to a feature map through successive residual blocks and then decoded with up-sampling layers to reconstruct the image. Each block contains a convolutional layers with nonlinear activation, connected by identity skip connections. 
\begin{equation}
    \hat{f} = D([V_{af}, V_{id}])
\end{equation}

\subsubsection{Denoising Auto‑encoder}
While the $V_{af} \in R^{128}$ for facial affect enables fine‑grained reconstruction of facial images, its high dimensionality poses a substantial challenge for efficient exploration by the sampling algorithm. To derive a low-dimensional latent space for sampling, a denoising auto-encoder is used to reconstruct the 128-dimensional features produced by the facial affect branch after training, further compressing the affect-related representations $R_{encoder}(V_{af}) \in R^{16}$:
\begin{equation}
    \hat{V}_{af} = R_{decoder}(R_{encoder}(V_{af}))
\end{equation}
Details for $R$ are in Appendix \cref{DistFace}.

\subsection{Training Regime}
\subsubsection{Training dataset}
The model was trained on the BU-4DFE dataset \cite{zhang2008high}. The dataset provides high-resolution dynamic facial expression captured as video frames. For each subject, six frame sequences (about 100 frames in each) were captured showing facial expressions transition from neutral to one of the six prototypic facial expressions (anger, disgust, happy, fear, sad, and surprise), respectively. The database contains 101 subjects in total, with balanced genders and a variety of ethnic backgrounds.

\subsubsection{Loss functions}
The primary objective of the model is to reconstruct a face whose facial expression is determined solely by the affect branch and whose identity is determined solely by the identity branch of the encoder, with the two sources of information contributing independently. 

In the pretraining stage, the model aim to reconstruct the input samples $F$ by minimizing:
\begin{equation}
    \mathcal{L}_{rec} = \|f - \hat{f}\|_2
\end{equation}

In the rectifying stage, the encoder module $E$ was fixed. Two randomly sampled faces ($f_{af}$ and $f_{id}$), were provided respectively to the encoder, typically differing in both affect and identity, with each face routed to its corresponding branch to get respective embeddings:
\begin{equation}
\begin{split}
    v^f_{af} = &[E_{fe}(f_{af}), E_{au}(f_{af})] \\
    &v^f_{id} = E_{id}(f_{id})
\end{split}
\end{equation}

These two embeddings from different faces then went through \textit{Equations (2, 3)} to get a unique reconstructed face image $\hat{f}$. To enforce identity coherence between $f_{id}$ and $\hat{f}$, the model minimized an identity‑preservation loss:
\begin{equation}
    \mathcal{L}_{id} = \|v^f_{id} - E_{id}(\hat{f})\|_2
\end{equation}

Likely, the model minimized an affect‑preservation loss in parallel:
\begin{equation}
    \mathcal{L}_{af} = \|v^f_{af} - E_{af}(\hat{f})\|_2
\end{equation}
The model was trained for 50 epochs during the pretraining stage and 100 epochs during the rectifying stage, using a batch size of 32 and learning rates of 0.001 and 0.0001 for the respective stages.

\subsection{Generative Performance}
\begin{figure}[ht]
  \vskip 0.2in
  \begin{center}
    \centerline{\includegraphics[width=0.8\columnwidth]{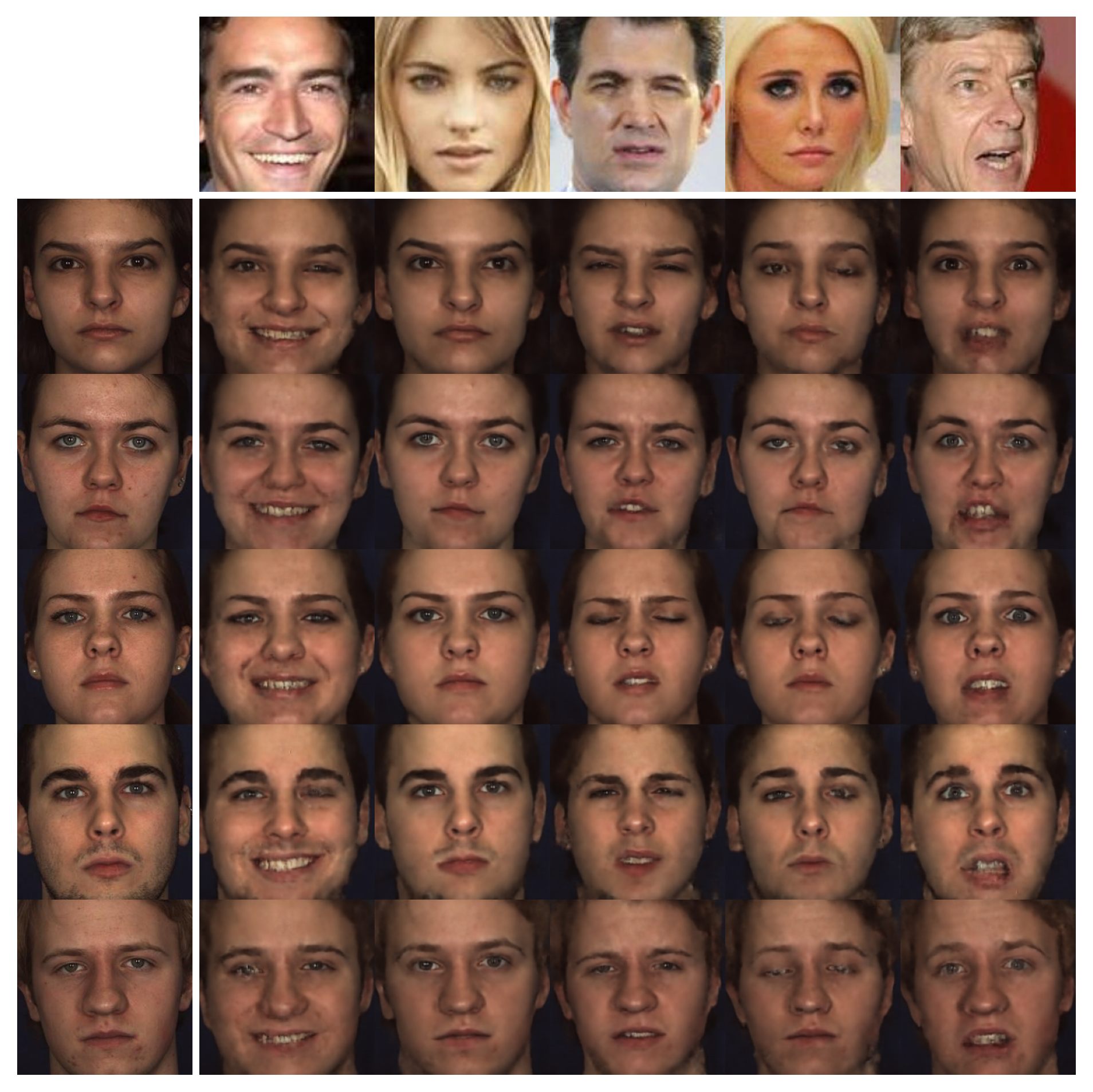}}
    \caption{
      Generative performance of our model on the affect‑transfer task. The right column shows randomly sampled identities from the in‑distribution dataset (BU‑4DFE), and the top row shows target facial affects selected from an out‑of‑distribution dataset (CelebA). The remaining grid displays images reconstructed by our model, combining the identity from the corresponding right‑column exemplar with the facial affect from the corresponding top‑row exemplar. The performance indicates that the model has learned a facial affect representation space that is effectively disentangled from facial identity.
    }
    \label{generative_performance}
  \end{center}
\end{figure}

During inference, the affect and identity branches process different input images and generate a fused facial representation that combines the identity from one source face with the affect from another, referred to as affect transfer. The performance of the trained model on affect-transfer is visualized in \cref{generative_performance}. The model demonstrates reliable performance in transferring the affect from in‑the‑wild faces outside the training set (the top row) onto arbitrary target identities (the right column).

\section{PriorProbe}

Within the 16-dimensional pure affect space, our proposed PriorProbe was performed at the individual level. PriorProbe roots in block Metropolis–Hastings algorithm where a sub-group of variables are updated together as a block in each single move \cite{robert1999monte}. In PriorProbe, the the face stimulus $f$ and the emotion category $e$ are treated as two variable blocks, updated separately in two kinds of trials. In each MCMC iteration, one block is sampled while conditioning on the current value of the other. 

\subsection{Face Trial for Categorical Representation}
The face trial follows a previous application of MCMCP which aims to identify people's category distribution $P(f|e)$ \cite{martin_testing_2012}. In a face trial, participants are presented with a emotion category $e$ and two face options ($f, f'$), asked to select one that best represents the category. The face stimuli are parameterized by the 16-dimensional affect vector within the latent space and reconstructed by the DistFace, with $f$ representing the current state and $f'$ representing the proposal one. Assuming that people make choices by matching the probability aligning with the Barker acceptance function \cite{barker_monte_1965},
\begin{equation}
    \alpha(f', f) = \frac{P(f'|e)}{P(f|e)+P(f'|e)}
\end{equation}
this provides the stochasticity that drives the MCMC sampling. As such, the target distribution in this block is the category representation $P(f|e)$ given a symmetric proposal function $q(\cdot)$ in the detailed balance:
\begin{equation}
    P(f|e)\alpha(f', f)q(f'|f) = P(f'|e)\alpha(f, f')q(f|f')
\end{equation}

However, this approach forces the MCMC sampler to navigate the entire state space with autocorrelations, resulting in substantial inefficiency. To accelerate sampling by restricting proposals to the “hot region’’ of each category, PriorProbe replaces the symmetric proposal with an Independent Metropolis–Hastings (IMH) scheme, in which the proposal distribution is independent of the current state and instead defined by ResEmoteNet called the Gatekeeper $G_f(f|e)$:
\begin{equation}
    G_f(f|e) \propto P_{res}(e|f)
\end{equation}

As ResEmoteNet only produces $P_{res}(e|f)$, to get face samples, $G_f(f|e)$ was determined through a MCMC sampling process. Therefore, the Gatekeeper only proposes faces that are likely to be members of corresponding emotion category. The detailed balance then becomes:
\begin{equation}
    \pi(f|e)\alpha(f', f)G_f(f'|e) = \pi(f'|e)\alpha(f, f')G_f(f|e)
\end{equation}
with the target distribution:
\begin{equation}
\pi(f|e) \propto P(f|e)G_f(f|e)
\end{equation}

\subsection{Categorization Trial for Categorical Prior}
Following each face trial is the categorization trial, in which participants are presented with the selected face $f$ along with two category options ($e, e'$) and asked to choose the one that best describes $f$. Following the IMH scheme, the proposal $e'$ is sampled from the distribution generated by ResEmoteNet which we regard as the Gatekeeper for categorization:
\begin{equation}
    G_e(e|f) = P_{res}(e|f)
\end{equation}
If a different category $e'$ is proposed, the trial is presented to participants for a choice; if the proposal matches the current state $e$, the current category is accepted as a new sample and the chain proceeds directly to the next face trial. The target distribution then becomes:
\begin{equation}
    \pi(e|f) \propto P(e|f)G_e(e|f)
\end{equation}

\subsection{Reweighting to Recover Individual Representations}
\label{recover_prior}
Each iteration of PriorProbe produce a $\{f, e\}$ sample from a face trial and a categorization trial, and the resulting Markov chain eventually converges on the joint distribution $\pi(f, e)$ to individuals. The category samples $e$s can therefore be treated as draws from the joint prior $\pi(e)$. To recover individual prior $P(e)$ from $\pi(e)$, each category sample is reweighted by $1/G_e(e|f)$, where $G_e(e|f)$ denotes the category probability assigned by ResEmoteNet. 



\section{Experiment}
The experiment aims to recover individuals’ cognitive priors over the seven basic facial affects (happy, surprise, sad, angry, disgust, fear, and neutral) using PriorProbe. Building on these inferred priors, we then examined whether incorporating human‑derived priors into a deep neural network improves its ability to predict individual performance in a facial expression recognition task.

\subsection{Participants}
Fifty-five participants were recruited through the Prolific online recruitment platform for this experiment. To ensure data quality, we applied the following screening criteria: native English speakers, normal vision, and a minimum of 100 prior submissions with an approval rate above 99\%. Participants received a compensation at a rate of £9.00/h. The average completion time was 1.5 hours for the experiment. This study was approved by the Department Research Ethics Committee at the University.

\subsection{Procedure}
\subsubsection{Recovering Priors} 
Participants went through PriorProbe procedure at first to recover their categorical priors. Each participant completed one thousand successive trials in this part. On each trial, participants either chose an affect category for a single face (categorization trial) or chose one of two faces that best matched a given category (face trial), with the two kinds of trials alternating randomly due to the stochastic nature of a Markov Chain. The facial identity of the stimuli on each trial was randomly drawn from a manually curated, gender‑balanced identity pool, and in every face trial the paired stimuli presented shared the same facial identity. 

Each participant completed seven block‑wise chains, with each chain initialized from one of the seven basic emotions. For example, chain $MC_{i1}$ denotes the first chain for participant $i$, which began with a categorization trial in which the face stimulus was sampled from the Gatekeeper's happy distribution $G_f(f|e=happy)$ and the response options were happy and one additional category selected at random. Thus, this design ensured that every participant navigated through the full categorical space during sampling.

\subsubsection{Classification of Novel and Ambiguous Stimuli} 
After completing the first part of the experiment, participants performed a separate facial affect categorization task. On each trial, they selected one of the seven basic affect categories that best described a presented face. The stimulus set consisted of 65 held-out real faces drawn from the CelebFaces Attributes Dataset (CelebA), a large‑scale collection of more than 200,000 celebrity images. To maximize sensitivity to individual differences and to provide a strong challenge for the deep neural network model (ResEmoteNet) in predicting individual choices, we selected only those images for which ResEmoteNet assigned a maximum category probability below 0.25. These low‑confidence images were among the most confusable for the model. To capture human-perceived uncertainty in each categorization trial, participants were also asked to report their confidence immediately after making each choice, on a scale from 1 (not confident at all) to 7 (extremely confident).

\subsection{Results}

\subsubsection{Individual differences among recovered priors}
\begin{figure}[ht]
  \vskip 0.2in
  \begin{center}
    \centerline{\includegraphics[width=0.6\columnwidth]{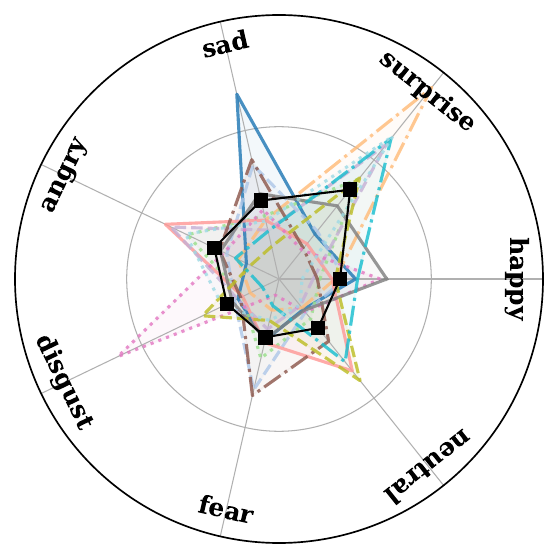}}
    \caption{
      Example individual priors recovered by PriorProbe. To preserve visual clarity and distinguishability across participants, we display priors from twelve randomly selected individuals. The black heptagon represents the average prior across individuals.
    }
    \label{individual_priors}
  \end{center}
\end{figure}

All individuals' priors over categories (i.e., $P_i(e)$) were recovered from participants' choices based on the procedures described in \cref{recover_prior}. In \cref{individual_priors}, we visualize the priors from a random subset of participants, as well as the average prior across all individuals. The heterogeneity among these shapes indicates substantial individual differences in priors over emotion categories. See all individual priors in Appendix \cref{recovered_priors}.

\subsubsection{Performance of prior-informed model}
\textbf{Model comparison.}
To examine whether the recovered individual priors are able to inform meaningful individual differences to the neural network trained on ground-truth of facial expression, we compared different models in terms of their predictive performance on individual decisions in the novel stimulus set. We directly combine these priors with the neural network within the inference space given their natural compatibility. The included models are listed in \cref{models}. For the ecological prior that reflects the environment people adapt to, we combined two large in-the-wild face datasets with emotion labels, RAF-DB \cite{li_reliable_2017} and FER2013, yielding a total of 51,216 faces and weighting each dataset by its respective sample size. The average prior was obtained by averaging bwMCMCPG-recovered priors across individuals. When predict individual decisions, the models containing individual-level priors (e.g., Individual prior) used the prior specific to the target participant.

\begin{table}[t]
  \caption{Included models for predicting individual classification.}
  \label{models}
  \begin{center}
    \begin{small}
        \begin{tabular}{ll}
          \toprule
          Model  & Expression   \\
          \midrule
          ResEmoteNet           & $P_{res}(e|f)$ \\
          Ecological prior         & $P_{eco}(e)$ \\
          Ecological prior + ResEmoteNet        & $P_{eco}(e)P_{res}(e|f)$ \\
          Average prior         & $\bar{P}(e)=\frac{1}{N}\sum_1^N{P_i(e)}$ \\
          Average prior + ResEmoteNet        & $\bar{P}(e)P_{res}(e|f)$ \\
          Individual prior     & $P_i(e)$ \\
          Individual prior + ResEmoteNet         & $P_i(e)P_{res}(e|f)$ \\
          \bottomrule
        \end{tabular}
    \end{small}
  \end{center}
  \vskip -0.1in
\end{table}

\begin{figure}[ht]
  \vskip 0.2in
  \begin{center}
    \centerline{\includegraphics[width=\columnwidth]{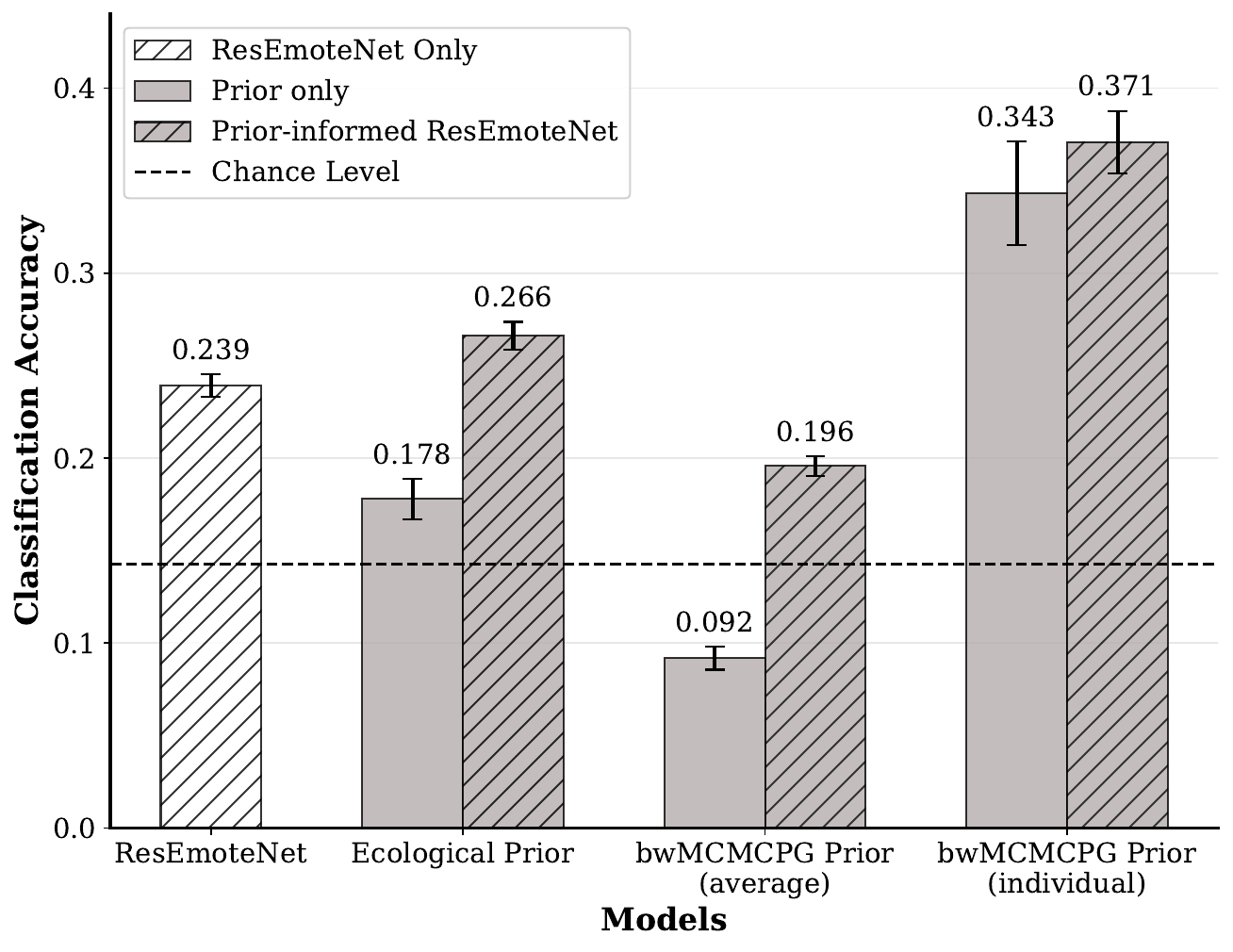}}
    \caption{
      Predictive performance on individual‑level categorization across models. ResEmoteNet is the neural network with SOTA performance in face expression recognition. The individual priors recovered by PriorProbe show the best performance in helping ResEmoteNet capture individual‑level responses to ambiguous stimuli.
    }
    \label{model_comparison}
  \end{center}
\end{figure}

The predictive performance of the models on individual decisions are presented in \cref{model_comparison}. ResEmoteNet achieves an average accuracy of 0.239 (SD=0.046) that is above chance (0.143). Among the three prior-alone model, only the PriorProbe‑recovered individual priors yield performance significant better than ResEmoteNet (t=3.743, p$<$.001), indicating that individual priors alone can attain higher alignments with individual observers than a state‑of‑the‑art neural network trained on large‑scale datasets. When these priors are integrated with ResEmoteNet, all three lead to performance gains relative to their corresponding pure‑prior models; individual priors recovered by PriorProbe yield the largest performance improvements over ResEmoteNet (t=8.544, p$<$.001), raising accuracy to 0.371 (SD=0.125). These results demonstrate that PriorProbe offers a robust signal for adapting neural networks to individual‑level preferences. 

\begin{figure*}[ht]
  \vskip 0.2in
  \begin{center}
    \centerline{\includegraphics[width=1.49\columnwidth]{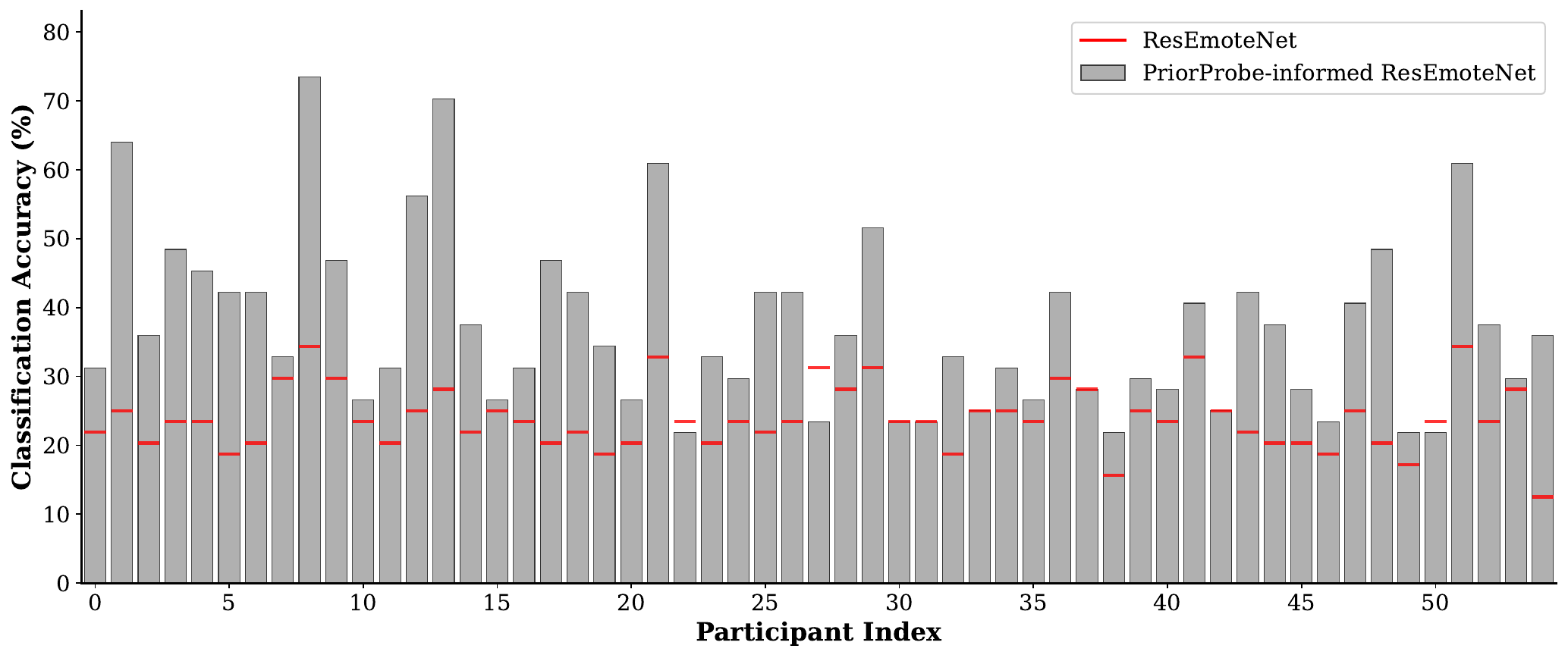}}
    \caption{
      Predictive performance of ResEmoteNet (red line) and PriorProbe-informed ResEmoteNet (grey bars) on individual‑level classification of ambiguous faces across all participants.
    }
    \label{individual_accuracies}

    \vspace{0.3cm}

    \centerline{\includegraphics[width=1.49\columnwidth]{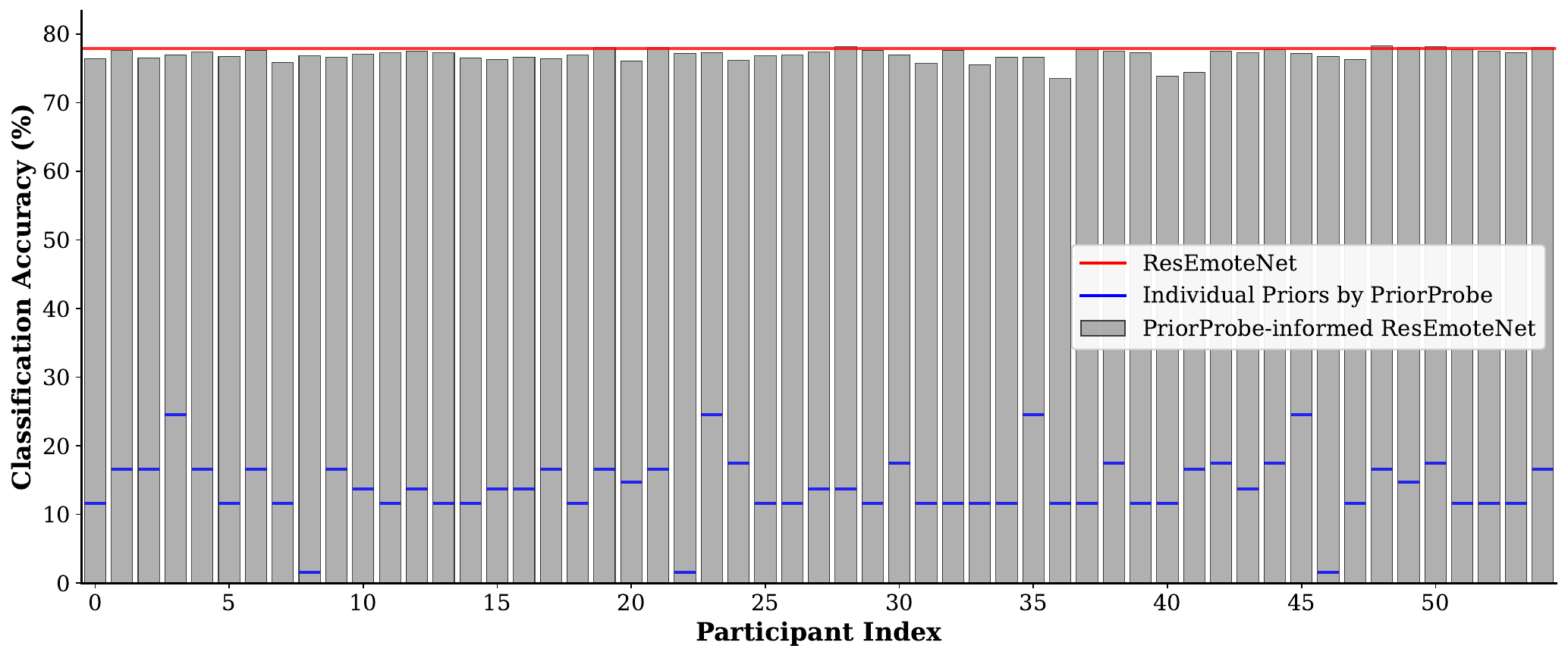}}
    \caption{
      Predictive performance of ResEmoteNet (red line), Prior-only (blue line) and PriorProbe-informed ResEmoteNet (grey bars) on ground-truth facial expression recognition, using test set from FER2013. 
    }
    \label{fer2013_individual_accuracies}
  \end{center}
\end{figure*}

\textbf{Individual-level performance.}
We further checked the performance gain at the individual level in \cref{individual_accuracies}. Across the 55 individual classification results, individual-prior-informed ResEmoteNet presents substantial improvement over ResEmoteNet in 49 of them. Overall, the recovered individual priors show great potential of informing neural networks with individual-level preference, particularly for ambiguous stimuli.

\begin{figure}[ht]
  \vskip 0.2in
  \begin{center}
    \centerline{\includegraphics[width=\columnwidth]{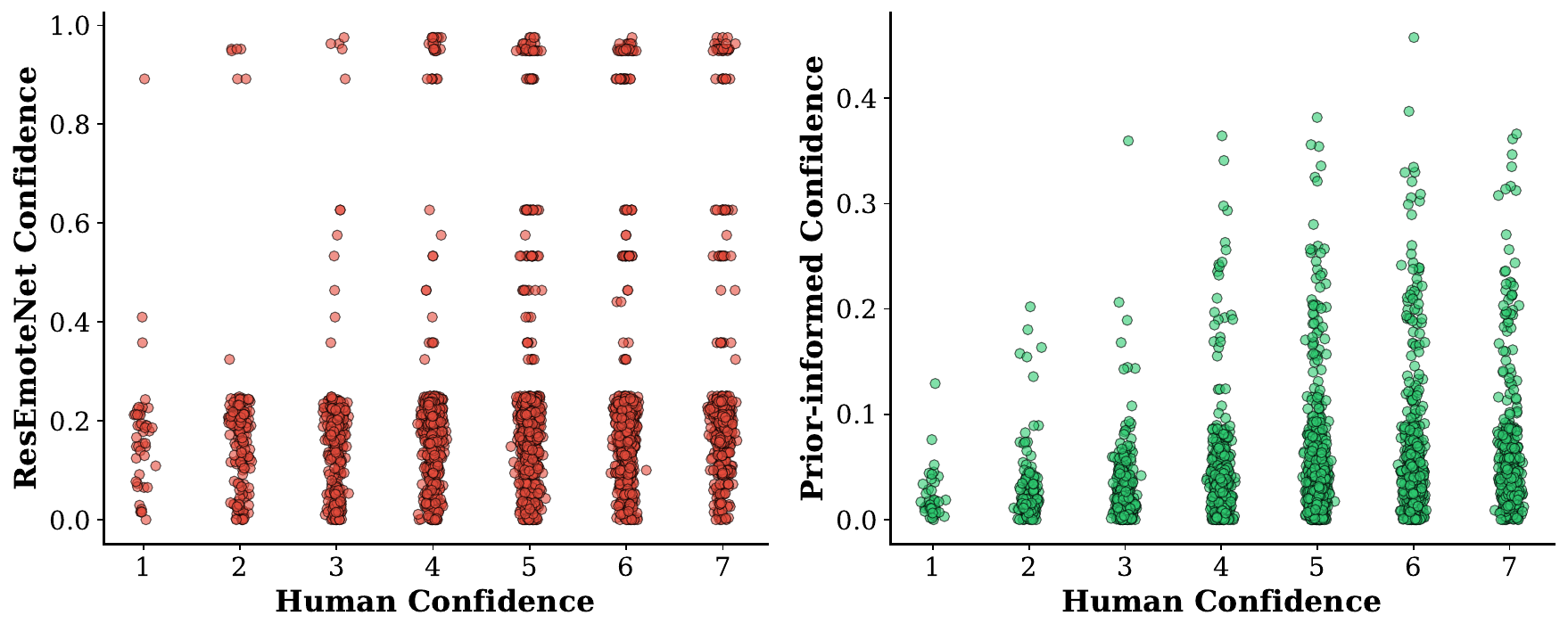}}
    \caption{
      Correlation between the probability assigned to correct label on each face by ResEmoteNet and the confidence in each trial rated by human participants. (Left) ResEmoteNet alone; (right) PriorProbe-informed ResEmoteNet.
    }
    \label{model_confidence}
  \end{center}
\end{figure}

\textbf{Improvement on confidence correlation.}
Beyond predicting individual classification of facial expression, we also checked the more fine-grained correspondence between the confidence of each decision produced by the prior-informed ResEmoteNet (assigned probability to correct labels) and rated by human participants. As shown in \cref{model_confidence}, after informed by individual priors, the correlation between confidence of ResEmoteNet and humans yields an improvement from 0.119 to 0.167. 

\textbf{Sensitivity check.}
So far we have merely examined the performance of the prior-informed model on predicting individual classifications of ambiguous faces. However, integrating individual priors explicitly with neural networks may undermine their original performance on predicting ground-truth labels of natural stimuli, even though they were proven to be more aligned with individuals on predicting ambiguous stimuli. To check this sensitivity, we tested the PriorProbe-informed ResEmoteNet on the test set from FER2013, which contains 3589 face images across 7 basic emotions. We also included the individual priors only (blue lines in \cref{fer2013_individual_accuracies}). The results are shown in \cref{fer2013_individual_accuracies}, no apparent decline of performance on ResEmoteNet is observed across all individuals when combining individual priors (mean accuracy=76.93\%). However, when relying on the priors only, the mean accuracy drops to 13.99\%, even though these priors achieve comparable performance with prior-informed ResEmoteNet when predicting classification of ambiguous stimuli (see \cref{model_comparison}). The result supports that individual priors recovered by PriorProbe improve individual alignment of ResEmoteNet without interfering its inference on ground-truth. 


\section{Conclusion}

In this study we present PriorProbe, a general method to jointly identify individual-specific priors over categories along with category representations. Meanwhile, we design DistFace for constructing an expressive sample space for PriorProbe in our case. By integrating the recovered priors with a state‑of‑the‑art deep neural network within inference space, we observed substantial gains for the prior-informed neural network in predicting individual classification of ambiguous stimuli, surpassing both ecological priors and population‑average priors. More strikingly, the integration does not interfere the neural network's inference on ground-truth labels. Our experiments demonstrate that the individual‑specific priors recovered by PriorProbe constitute a dependable basis for informing neural networks about personal preference structure.

\section*{Acknowledgements}

This work and related results were made possible with the support of H. Yan's Chancellor’s International Scholarship from the University of Warwick.


\section*{Impact Statement}

This paper represents an effort to advance the personalization of deep neural networks. We hope that our approach marks a step toward individual‑aware AI systems that are more helpful, adaptive, and applicable to human needs.

\bibliography{references}
\bibliographystyle{icml2026}

\newpage
\appendix
\onecolumn
\section{DistFace Architecture}
\label{DistFace}

\begin{figure}[ht]
  \vskip 0.2in
  \begin{center}
    \centerline{\includegraphics[width=0.6\columnwidth, trim={9cm 2cm 9cm 2cm}, clip]{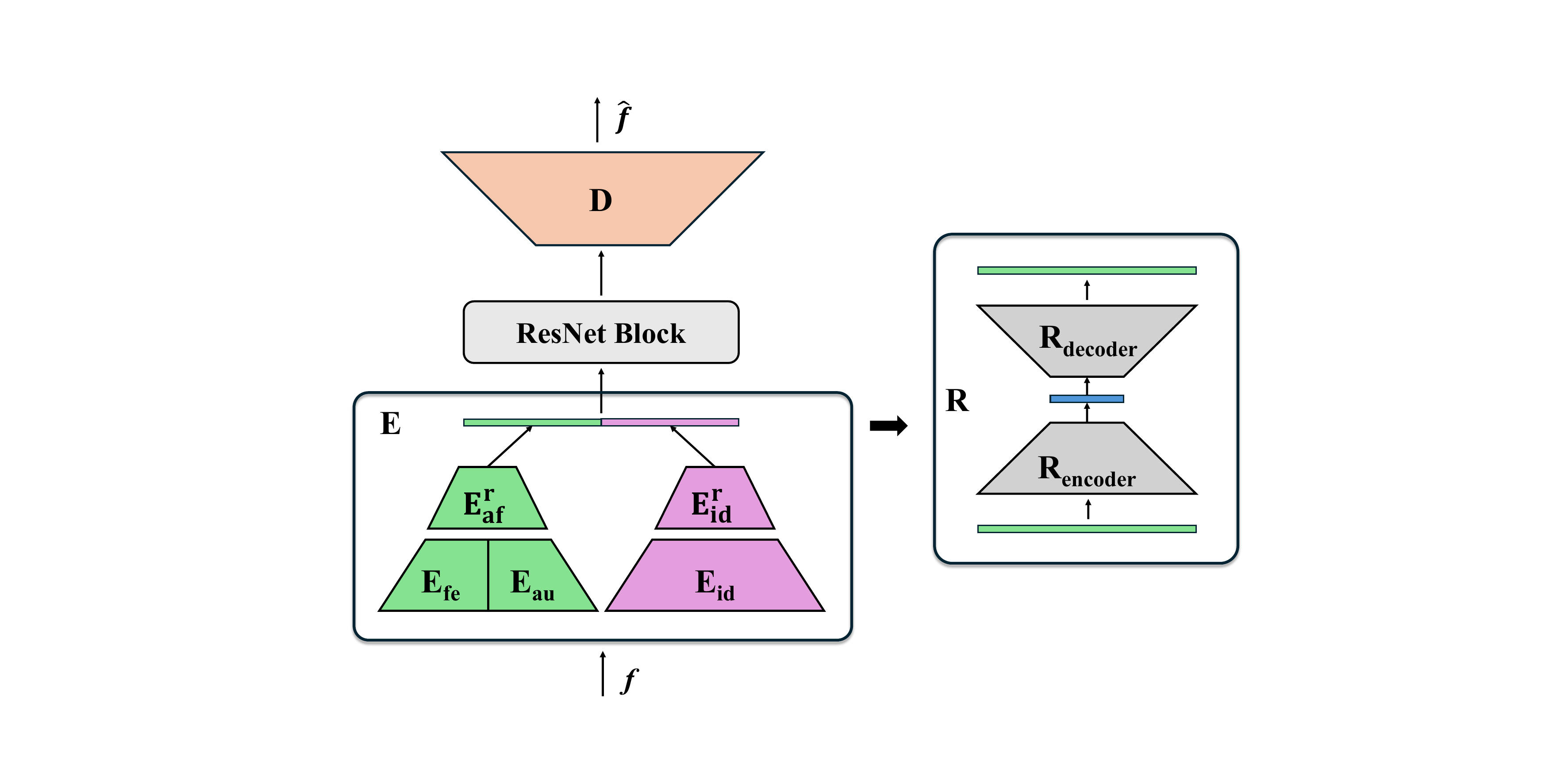}}
    \caption{
      Structure of face generator and embedding reduction model $R$ in proposed DistFace. The affect and identity branches work independently within the encoder module, while merging and synthesizing new images in the decoder $D$. The auto-encoder $R$ reconstruct the affect embeddings and compress them into a smaller latent space, for efficient sampling by PriorProbe. 
    }
    \label{model_arc}
  \end{center}
\end{figure}

Structure of DistFace is shown in \cref{model_arc}. Specifically, the auto-encoder $R$ was trained after training the generation model. The affect embeddings of BU4D dataset produced by the generation model was taken as training data for $R$. Through reconstructing these embeddings, $R$ captures low-dimensional representations of facial affect, serving as the basis for PriorProbe.

\newpage
\section{All Individual Priors Recovered by PriorProbe}
\label{recovered_priors}

\begin{figure}[ht]
  \vskip 0.2in
  \begin{center}
    \centerline{\includegraphics[width=0.5\columnwidth]{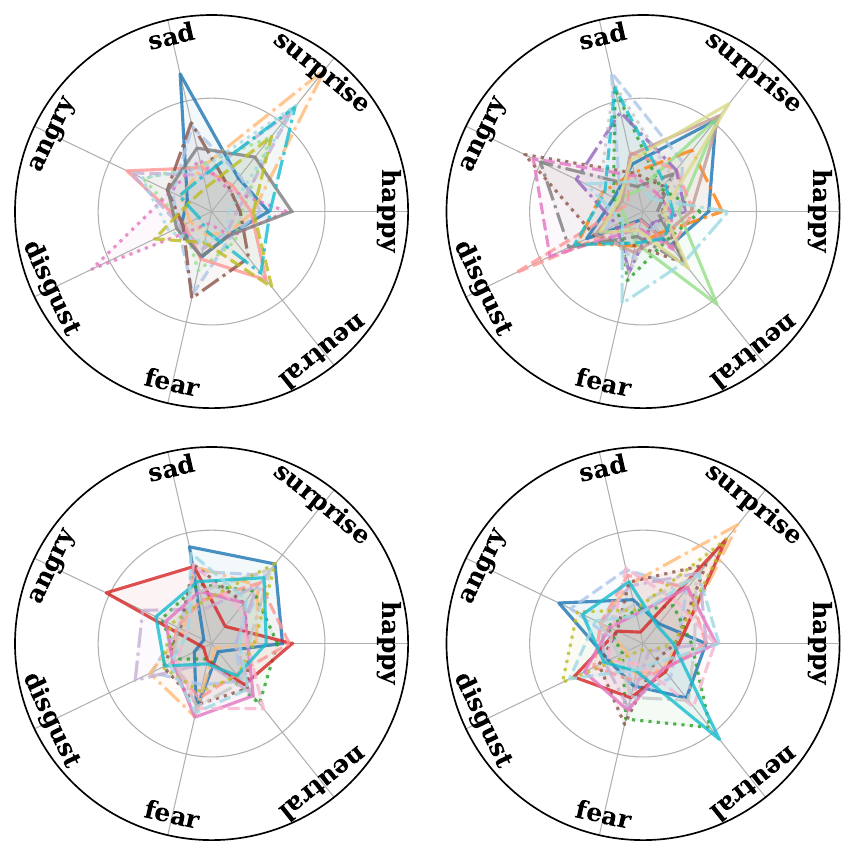}}
    \caption{
      PriorProbe-recovered individual priors across 55 participants. 
    }
    \label{all_priors}
  \end{center}
\end{figure}

\section{Further Discussion and Limitations}

Individual priors identified by bwMCMCPG exhibit substantial individual differences, providing a principled way to model individual differences in neural networks. To our knowledge, individual variability in neural networks stems either from architectural and initialization factors (e.g., random seeds, different activation functions) \cite{mehrer_individual_2020} or from exposure to distinct training datasets. These sources of variation do not guarantee that the resulting individual specifications will align with individual behavioral patterns. Our proposed bwMCMCPG therefore shows strong potential as a general method for inducing human‑aligned individual differences in neural networks---any cognitive priors involved in inductive inference can, in principle, be formulated as decision‑making trials within bwMCMCPG and recovered accordingly. The recovered individual priors can then support a wide range of downstream applications: they not only enhance individual‑level modeling in neural networks but also enable the incorporation of personal preference structures into agent‑based populations, facilitating the study in social behavior.

Furthermore, compared with collecting individualized training data, recovering structured mental representations offers a more fundamental route to personalizing modern AI models. For example, contemporary large language models (LLMs) typically infer information about an individual through in‑context learning, drawing on system instructions and the user’s own conversational history. Although these foundation models demonstrate impressive capacity to infer user personas from highly sparse data \cite{cheng_binding_2023}, they are highly sensitive to subtle contextual variations, which users may not articulate consistently or accurately. Accordingly, endowing LLMs with explicit individual representations tailored to the relevant context is likely to yield more stable personalization than relying on limited contextual cues alone.

Beyond the priors, bwMCMCPG also recovers how individuals represent categories as well (i.e., the likelihood function in our categorization task), which provides additional fine-grained information on individual preferences \cite{yan_quickly_2024}. These categorical representations can also be incorporated directly into a network’s decision process, enabling the model to capture individual behavior more comprehensively. Moreover, given a likelihood function that specifies the distribution over real‑world stimuli conditioned on a hypothesis, the recovered priors and likelihoods can serve as individualized generative models. This allows us to generate individualized training data that neural networks can use to internalize these representations, rather than relying on the multiplicative augmentation strategy used here---a more thorough integration.

A key limitation of the present work lies in the restricted size of the individual decision‑making task (65 face images). Increasing the experimental scale in future studies will be important for establishing more robust evidence of the performance improvements attributed to bwMCMCPG. In addition, the present analysis  did not take within‑subject variability in the decision‑making task into consideration. Actually, individuals often respond inconsistently across repeated trials, and incorporating this variability would allow for more accurate calibration of individual‑level models. Finally, it takes 1.5 ours in this case to extract individual-level prior, which means the method is more applicable for critical decision situations, rather than everyday judgments. Future work should validate the method in other critical domains to establish its generality and broader applicability. 

\end{document}